\documentclass[11pt,letterpaper]{article}
\usepackage{emnlp2016}
\usepackage{times}
\usepackage{latexsym}
\usepackage{amsfonts}
\usepackage{graphicx}
\usepackage{bm}
\usepackage{amsmath}
\usepackage{multirow}
\usepackage{eqlist}
\usepackage{stmaryrd}
\usepackage{subfig}
\usepackage{enumitem}
\def\emnlppaperid{***}
\makeatletter
\DeclareRobustCommand{\cev}[1]{%
  \mathpalette\do@cev{#1}%
}
\newcommand{\do@cev}[2]{%
  \fix@cev{#1}{+}%
  \reflectbox{$\m@th#1\vec{\reflectbox{$\fix@cev{#1}{-}\m@th#1#2\fix@cev{#1}{+}$}}$}%
  \fix@cev{#1}{-}%
}
\newcommand{\fix@cev}[2]{%
  \ifx#1\displaystyle
    \mkern#23mu
  \else
    \ifx#1\textstyle
      \mkern#23mu
    \else
      \ifx#1\scriptstyle
        \mkern#22mu
      \else
        \mkern#22mu
      \fi
    \fi
  \fi
}
\makeatother

\title{Recognizing Implicit Discourse Relations via Repeated Reading: \\Neural Networks with Multi-Level Attention}
\author{Yang Liu$^{1,2}$, Sujian Li$^1$\\
$^1$ Key Laboratory of Computational Linguistics, Peking University, MOE, China\\
$^2$ ILCC, School of Informatics, University of Edinburgh, United Kingdom\\
  {\tt \{cs-ly, lisujian\}@pku.edu.cn}}
\date{}
\begin{document}
\maketitle
\begin{abstract}
 Recognizing implicit discourse relations is a challenging but important task in the field of Natural Language
Processing.
For such a complex text processing task, different from previous studies, we argue that it is necessary to
repeatedly read the arguments  and dynamically exploit the efficient features useful for
recognizing discourse relations.
To mimic the repeated reading strategy, we propose the neural networks with multi-level attention (NNMA),
combining the attention mechanism and  external memories to gradually fix the attention on some specific words  helpful to judging the discourse relations.
Experiments on the PDTB dataset show that our proposed method achieves the state-of-art results.
The visualization of the attention weights also illustrates the progress that our model  observes the arguments on  each level and progressively locates the important words.
\end{abstract}
\section{Introduction}
Discourse relations (e.g., contrast and causality) support a set of sentences to form a coherent text. Automatically recognizing discourse relations can help many downstream tasks such as question answering and automatic summarization.
Despite great progress in classifying explicit discourse relations  where the discourse connectives (e.g., ``because'', ``but'') explicitly exist in the text, implicit discourse relation recognition  remains a challenge due to the absence of discourse connectives.
Previous research mainly focus on exploring various kinds of efficient features and machine learning models to classify the implicit discourse relations~\cite{soricut2003sentence,baldridge2005probabilistic,subba2009effective,hernault2010hilda,pitler2009automatic,joty2012novel}. 
To some extent, these methods  simulate the single-pass  reading process that a person quickly skim the text through one-pass reading and directly collect important clues for understanding the text.
Although single-pass reading plays a crucial role  when we just want the general meaning and do not necessarily need to understand every single point of the text, it is not enough for tackling tasks that need a deep analysis of the text.
In contrast with single-pass reading, repeated reading involves the process where learners repeatedly read the text in detail with specific learning aims, and has the potential to improve readers' reading fluency and comprehension of the text (National Institute of Child Health and Human Development, 2000; LaBerge and Samuels, 1974).
Therefore, for the  task of discourse parsing, repeated reading is necessary, as it is difficult to generalize which words are really useful on the first try  and efficient features should be dynamically exploited through several passes of reading .

Now, let us check one real example to elaborate the necessity of using repeated reading in discourse parsing.
\begin{description}
\vspace{-1ex}
\item[Arg-1]:
the use of 900 toll numbers has been \textit{expanding rapidly} in recent years
\item[Arg-2]:
for a while, high-cost pornography lines and services
that tempt children to dial (and redial) movie or music
information \textit{earned the service a somewhat sleazy image}
\\
\hspace*{\fill}{(Comparison - wsj\_2100)}
\end{description}

To identify the ``\textbf{Comparison}'' relation between the two arguments \textit{Arg-1} and \textit{Arg-2}, the most crucial clues mainly lie in some content, like \textit{``expanding rapidly''} in \textit{Arg-1} and \textit{``earned the service a somewhat sleazy image''} in \textit{Arg-2}, since there exists a contrast between  the semantic meanings of these two text spans.
However, it is difficult to obtain sufficient information for pinpointing these words through scanning the argument pair left to right in one pass.
In such case, we  follow the repeated reading strategy, where we obtain the general meaning through reading the arguments for the first time, re-read them later and gradually pay close attention to the key content.

Recently, some approaches simulating repeated reading have witnessed their success in different tasks. These models  mostly combine the attention mechanism that has been originally designed to solve the alignment problem in machine translation~\cite{bahdanau2014neural} and the external memory which can be read and written when processing the objects~\cite{sukhbaatar2015end}.  
For example, \newcite{DBLP:journals/corr/KumarISBEPOGS15} drew attention to specific facts of the input sequence and processed the sequence via multiple hops to generate an answer.
In computation vision, \newcite{DBLP:journals/corr/YangHGDS15} pointed out that repeatedly giving attention to different regions of an image  could gradually lead to  more precise image representations.

Inspired by these recent work, for discourse parsing, we propose a model that aims to repeatedly read an argument pair  and gradually focus on more fine-grained parts after grasping the global information.
Specifically, we design the Neural Networks with Multi-Level Attention (NNMA) consisting of one general level and several attention levels.
In the general level, we capture the general representations of each argument based on two bidirectional long short-term memory (LSTM) models.
For each attention level, NNMA generates a weight vector over the argument pair to locate the important parts related to the discourse relation.
And an external short-term memory is designed to store the information exploited in previous levels and help update the argument representations.
We stack this structure in a recurrent manner, mimicking the process of reading the arguments multiple times.
Finally, we use the representation output from the highest attention level to identify the discourse relation.
Experiments on the PDTB dataset show that our proposed  model  achieves the state-of-art results.

\section{Repeated Reading Neural Network with Multi-Level Attention}
In this section, we describe how we use the neural networks with multi-level attention to repeatedly read the argument pairs and recognize implicit discourse relations.

\begin{figure}[!htbp]
  \centering
  \includegraphics[width=3in]{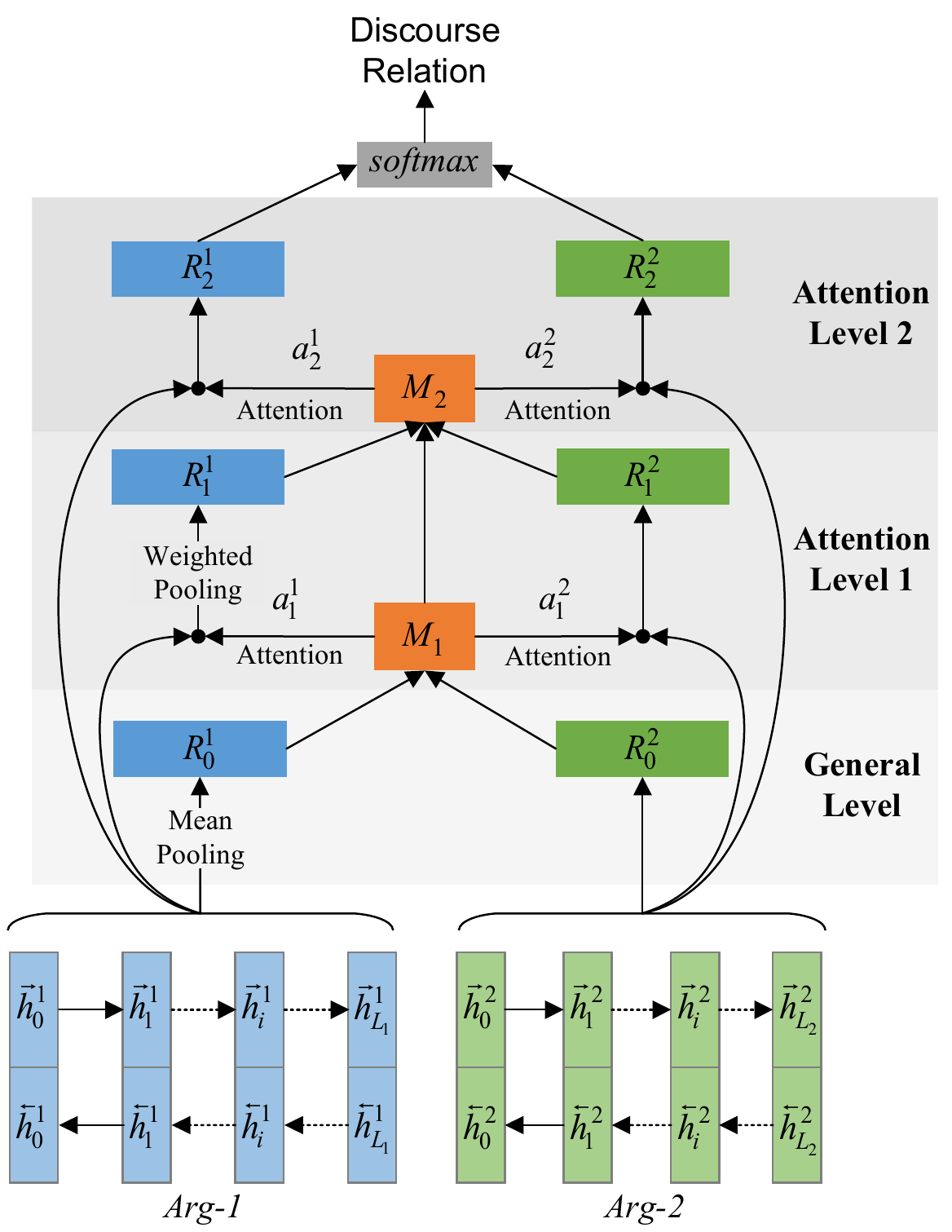}
  \caption{ Neural Network with Multi-Level Attention. (Two attention levels are given here.) }
  \label{fig:model}
\end{figure}

 First, we get the general understanding of the arguments through skimming them.
 To implement this, we adopt the bidirectional Long-Short Term Memory Neural Network (bi-LSTM)  to model each argument, as bi-LSTM is good at modeling over a  sequence of words and can represent each word with consideration of more contextual information.
Then, several attention levels are designed to simulate the subsequent multiple passes of reading.
On each attention level, an external short-term memory is used to store what has been learned from previous passes and guide  which words should be focused on. 
To  pinpoint the useful parts of the arguments, the attention mechanism is used to predict a probability distribution over each word, indicating to what degree each word should be concerned.
The overall architecture of our model is shown in Figure~\ref{fig:model}.
For clarity, we only illustrate two attention levels in the figure. It is noted that we can easily extend our model to more attention levels.
\subsection{Representing Arguments with LSTM}
The Long-Short Term Memory (LSTM) Neural Network  is a variant of the Recurrent Neural Network which is usually used for modeling a sequence.
In our model, we adopt two LSTM neural networks to respectively model the two arguments: the left argument \textit{Arg-1} and the right argument \textit{Arg-2}.

First of all, we associate each word $w$ in our vocabulary with a vector representation $\bm{x}_w \in \mathbb{R}^{D_e}$.
Here we adopt the pre-trained vectors  provided by GloVe~\cite{pennington2014glove}.
Since an argument can be viewed as a sequence of word vectors, let $\bm{x}_i^1$ ($\bm{x}_i^2$) be the $i$-th word vector in  argument \textit{Arg-1} (\textit{Arg-2}) and
the two arguments can be represented as,
\begin{gather*}
Arg\text{-}1: [\bm{x}^1_1, \bm{x}^1_2, \cdots, \bm{x}^1_{L_1}]\\
Arg\text{-}2: [\bm{x}^2_1, \bm{x}^2_2, \cdots, \bm{x}^2_{L_2}]
\end{gather*}
where \textit{Arg-1} has $L_1$ words and \textit{Arg-2} has $L_2$ words.

To model the two arguments, we briefly introduce the working process how the LSTM neural networks model a sequence of words.
For the $i$-th time step, the model reads the $i$-th word $\bm{x}_i$ as the input and updates the output vector $\bm{h}_i$  as follows~\cite{DBLP:journals/corr/ZarembaS14}.
\begin{align}
\bm{i}_i =&~sigmoid(\bm{W}_i[\bm{x}_i,\bm{h}_{i-1}]+\bm{b}_i)\\
\bm{f}_i =&~sigmoid(\bm{W}_f[\bm{x}_i,\bm{h}_{i-1}]+\bm{b}_f)\\
\bm{o}_i =&~sigmoid(\bm{W}_o[\bm{x}_i,\bm{h}_{i-1}]+\bm{b}_o)\\
\tilde{\bm{c}}_i =&~tanh(\bm{W}_c[\bm{x}_i,\bm{h}_{i-1}]+\bm{b}_c)\\
\bm{c}_i =&~\bm{i}_i*\tilde{\bm{c}}_i+\bm{f}_i*\bm{c}_{i-1}\\
\bm{h}_i =&~\bm{o}_i*tanh(\bm{c}_i)
\end{align}
where $[ \quad ]$ means the concatenation operation of several vectors.  $\bm{i}, \bm{f}, \bm{o}$  and $\bm{c}$ denote the input gate, forget gate, output gate and memory cell respectively in the LSTM architecture.
The input gate $\bm{i}$ determines how much the input $\bm{x}_i$ updates the memory cell. The output gate $\bm{o}$ controls how much the memory cell influences  the output.
The forget gate $\bm{f}$ controls how the past memory $\bm{c}_{i-1}$ affects the current state.
$\bm{W}_i, \bm{W}_f, \bm{W}_o, \bm{W}_c, \bm{b}_i, \bm{b}_f, \bm{b}_o, \bm{b}_c$ are the network parameters.

Referring to the work of~\newcite{wang2015long}, we implement the bidirectional version of LSTM neural network to model the argument sequence. Besides processing the  sequence in the forward direction, the bidirectional LSTM (bi-LSTM) neural network also  processes  it in the reverse direction. 
As shown in Figure \ref{fig:model},  using two bi-LSTM neural networks, we can obtain $\bm{h}^1_i = [\vec{\bm{h}}^1_i, \cev{\bm{h}}^1_i]$ for the $i$-th word in \textit{Arg-1} and $\bm{h}^2_i = [\vec{\bm{h}}^2_i, \cev{\bm{h}}^2_i]$ for the $i$-th word in \textit{Arg-2}, where $\vec{\bm{h}}^1_i, \vec{\bm{h}}^2_i \in \mathbb{R}^{d}$ and $\cev{\bm{h}}^1_i, \cev{\bm{h}}^2_i \in \mathbb{R}^{d}$ are the output vectors from two directions.

Next, to get the general-level representations of the arguments, we apply a mean pooling operation over the bi-LSTM outputs,
and obtain  two vectors $\bm{R}_0^1$ and $\bm{R}_0^2$, which can reflect the global information of the argument pair.
\begin{align}
\bm{R}_0^1 = \frac{1}{L_1} \sum_{i=0}^{L_1}{\bm{h}_i^1}\\
\bm{R}_0^2 = \frac{1}{L_2}\sum_{i=0}^{L_2}{\bm{h}_i^2}
\end{align}
\subsection{Tuning Attention via Repeated Reading}
After obtaining the general-level representations by treating each word equally, we simulate the repeated reading and design multiple attention levels to gradually pinpoint those words particularly useful for discourse relation recognition.
In each attention level, we adopt the attention mechanism to determine which words should be focused on.
An external short-term memory is designed to remember what has seen in the prior levels and guide the attention tuning process in current level. 

Specifically, in the first attention level,  we concatenate $\bm{R}_0^1$, $\bm{R}_0^2$ and $\bm{R}_0^1{\scalebox{0.75}[1.0]{\( - \)}}\bm{R}_0^2$ and apply a non-linear transformation over the concatenation to catch the general understanding of the argument pair. 
The use of $\bm{R}_0^1{\scalebox{0.75}[1.0]{\( - \)}}\bm{R}_0^2$  takes a cue from the difference between two vector representations which has been found explainable and meaningful in many applications~\cite{mikolov2013linguistic}.
Then, we get the memory vector $\bm{M}_1\in \mathbb{R}^{d_m}$ of the first attention level   as
\begin{align}
\bm{M}_1 = tanh(\bm{W}_{m,1}[\bm{R}_0^1, \bm{R}_0^2, \bm{R}_0^1{\scalebox{0.75}[1.0]{\( - \)}}\bm{R}_0^2])
\end{align}
where  $\bm{W}_{m,1} \in \mathbb{R}^{d_m\times 6d}$ is the weight matrix.

With $\bm{M}_1$ recording the general meaning of the argument pair, our model re-calculates the importance of each word.
We assign each word a weight measuring to what degree  our model should pay attention to it.
The weights are so-called  ``attention'' in our paper. 
This process is designed to simulate the process that we re-read the arguments and pay more attention to some specific words with an overall understanding derived from the first-pass reading.
Formally, for \textit{Arg-1}, we use the memory vector $\bm{M}_1$ to update the representation of each word with a non-linear transformation.
According to the updated word representations $\bm{o}^{1}_1$, we get the attention vector $\bm{a}^1_1$.
\begin{align}
\bm{h}^1 =&~[\bm{h}_0^1,\bm{h}_1^1,\cdots,\bm{h}_{L_1}^1]\\
\bm{o}^{1}_1 =&~tanh(\bm{W}_{a,1}^1\bm{h}^1+\bm{W}_{b,1}^1(\bm{M}_1\otimes \bm{e}))\\
\bm{a}^1_1 =&~softmax(\bm{W}_{s,1}^1\bm{o}^1_1)
\end{align}
where $\bm{h}^1 \in \mathbb{R}^{2d\times L_1}$ is the concatenation of all LSTM output vectors of \textit{Arg-1}.
$\bm{e}\in \mathbb{R}^{L_1}$ is a vector of $1$s and the $M_1 \otimes \bm{e}$ operation denotes that we repeat the vector $M_1$ $L_1$ times and generate a ${{d_m}\times L_1}$  matrix. %
The attention vector $\bm{a}^1_1 \in \mathbb{R}^{L_1}$ is obtained through applying a $softmax$ operation over $\bm{o}^1_1$.
${\bm{W}_{a,1}}^1\in \mathbb{R}^{2d\times2d}, {\bm{W}_{b,1}}^1\in \mathbb{R}^{2d\times d_m}$ and ${\bm{W}_{s,1}}^1 \in \mathbb{R}^{1\times2d}$  are the transformation weights. It is noted that the subscripts denote the current attention level and the superscripts denote the corresponding argument. 
In the same way, we can get the attention vector $\bm{a}^2_1$ for \textit{Arg-2}.

Then, according to  $\bm{a}_1^1$ and $\bm{a}_1^2$, our model re-reads the arguments  and get the new representations  $\bm{R}^1_1$ and $\bm{R}^2_1$ for the first attention level.
\begin{align}
\bm{R}^1_1 = \bm{h}^1(\bm{a}^1_1)^T\\
\bm{R}^2_1 = \bm{h}^2(\bm{a}^2_1)^T
\end{align}

Next, we iterate the  ``memory-attention-representation'' process and design more attention levels, giving NNMA the ability to gradually infer more precise attention vectors.
The processing of the second or above attention levels is slightly different from that of the first level, as we update the memory vector in a recurrent way.
To formalize, for the $k$-th attention level ($k \geq 2$), we use the following formulae for \textit{Arg-1}.
\begin{align}
\bm{M}_k =&~tanh(\bm{W}_{m,k}[\bm{R}_{k\scalebox{0.45}[1.0]{\(-\)}1}^1, \bm{R}_{k\scalebox{0.45}[1.0]{\(-\)}1}^2, \bm{R}_{k\scalebox{0.45}[1.0]{\(-\)}1}^1{\scalebox{0.75}[1.0]{\(-\)}}\bm{R}_{k\scalebox{0.45}[1.0]{\(-\)}1}^2,\bm{M}_{k\scalebox{0.45}[1.0]{\(-\)}1}])\\
\bm{o}^1_k =&~tanh(\bm{W}^1_{a,k}\bm{h}^1+\bm{W}^1_{b,k}(\bm{M}_k\otimes \bm{e}))\\
\bm{a}^1_k =&~softmax(\bm{W}^1_{s,k}\bm{o}^1_k)\\
\bm{R}_k^1 =&~\bm{h}^1(\bm{a}_k^1)^T
\end{align}
In the same way, we can computer $\bm{o}^2_k, \bm{a}^2_k$ and $\bm{R}^2_k$ for \textit{Arg-2}.

Finally, we use the newest representation derived from the top attention level to recognize the discourse relations.
Suppose there are totally $K$ attention levels and $n$ relation types, the predicted discourse relation distribution $\bm{P} \in \mathbb{R}^{n}$ is calculated as
\begin{align}
&\bm{P} = softmax(\bm{W}_p[\bm{R}_K^1,\bm{R}_K^2, \bm{R}^1_K{\scalebox{0.75}[1.0]{\( - \)}}\bm{R}_K^2]+ \bm{b}_p )
\end{align}
where $\bm{W}_p \in \mathbb{R}^{n\times6d}$ and  $\bm{b}_{p} \in \mathbb{R}^{n}$ are the transformation weights.
\subsection{Model Training}
To train our model, the training objective is defined as the cross-entropy loss between the outputs of the softmax layer and the ground-truth class labels.
We use  stochastic gradient descent (SGD) with momentum to train the neural networks.

To avoid over-fitting,  dropout operation is applied on the top feature vector before the softmax layer.
Also,   we  use different learning rates $\lambda$ and $\lambda_e$ to train the neural network parameters $\Theta$  and the word embeddings $\Theta_e$ referring to \cite{TACL536}. $\lambda_e$ is set to a small value  for preventing over-fitting on this task. In the experimental part, we will introduce the setting of the hyper-parameters.

\section{Experiments}
\subsection{Preparation}
We evaluate our model on the Penn Discourse Treebank (PDTB)~\cite{10532931}.
In our work, we experiment on the four top-level classes in this corpus as in previous work ~\cite{rutherfordimproving}.
We extract all the implicit relations of PDTB, and follow the setup of ~\cite{rutherfordimproving}. We split the data into a training set (Sections 2-20), development set (Sections 0-1), and test set (Section 21-22).
Table~\ref{tab:imp} summarizes the statistics of the four PDTB discourse relations, i.e., Comparison, Contingency, Expansion and Temporal.
\begin{table}[!htbp]
\center
  \begin{tabular}{|l|c|c|c|}
  \hline
Relation&Train&Dev&Test\\
\hline
Comparison&1855&189&145\\
Contingency&3235&281&273\\
Expansion&6673&638&538\\
Temporal&582&48&55\\
\hline
\textbf{Total}&12345&1156&1011\\
  \hline
\end{tabular}
\caption{Statistics of Implicit Discourse Relations in  PDTB.}
  \label{tab:imp}
\end{table}

We first convert the tokens in PDTB to lowercase. 
The word embeddings used for initializing the word representations are  provided by GloVe~\cite{pennington2014glove}, and the dimension of the embeddings $D_e$ is 50.  The hyper-parameters, including the momentum $\delta$, the two learning rates $\lambda$ and $\lambda_e$, the dropout rate $q$, the dimension of LSTM output vector $d$, the dimension of memory vector $d_m$ are all set according to the performance on the development set
Due to space limitation, we do not present the details of tuning the hyper-parameters and only give their final settings as shown in Table~\ref{tab:hyp}. 
\begin{table}[!htbp]
\center
  \begin{tabular}{|c|c|c|c|c|c|c|}
  \hline
$\delta$&$\lambda$&$\lambda_e$&$q$&$d$&$d_m$\\
  \hline
0.9&0.01&0.002&0.1&50&200\\
  \hline
\end{tabular}
\caption{Hyper-parameters for Neural Network with
Multi-Level Attention.}
  \label{tab:hyp}
\end{table}

To evaluate our model, we adopt two kinds of experiment settings. The first one is the four-way classification task, and the second one is the binary classification task, where we build a one-vs-other classifier for each class.
For the second setting, to solve the problem of unbalanced classes in the training data, 
we follow the reweighting method of~\cite{rutherfordimproving}  to reweigh the training instances according to the size of each relation class.
We also use visualization methods to analyze how multi-level attention helps our model.
\subsection{Results}
First, we design  experiments to evaluate the effectiveness of attention levels and how many attention levels are appropriate.
To this end, we implement a baseline model 
(LSTM with no attention) 
which directly applies the mean pooling operation over  LSTM output vectors of two arguments without any attention mechanism. 
Then we consider different attention levels including one-level, two-level and three-level.
The detailed results are shown in Table~\ref{tab:results1}.
For four-way classification, macro-averaged $F1$ and Accuracy are used as evaluation metrics. For binary classification, $F1$ is adopted to evaluate the performance on each class.
\begin{table}[!htbp]
\centering
\footnotesize
\begin{tabular}{|>{\centering\arraybackslash}p{4.5em}|>{\centering\arraybackslash}p{1.85em}>{\centering\arraybackslash}p{1.85em}|>{\centering\arraybackslash}p{1.85em}>{\centering\arraybackslash}p{1.85em}>{\centering\arraybackslash}p{1.85em}>{\centering\arraybackslash}p{1.85em}|}
\hline
\multirow{2}{*}{System}                                      & \multicolumn{2}{c|}{Four-way}   & \multicolumn{4}{c|}{Binary}   \\ \cline{2-7}
                                                             & $F1$             & Acc.           & Comp. & Cont. & Expa.  & Temp. \\ \hline
\begin{tabular}[c]{@{}c@{}} LSTM\end{tabular}       & 39.40          & 54.50          &33.72       &44.79       &68.74       &33.14       \\ \hline
\begin{tabular}[c]{@{}c@{}}NNMA\\ (one-level)\end{tabular}   & 43.48          & 55.59          & 34.72      &49.47       & 68.52      &36.70       \\ \hline
\begin{tabular}[c]{@{}c@{}}NNMA\\ (two-level)\end{tabular}   & \textbf{46.29} & 57.17                               & 36.70                 & \textbf{54.48}        & \textbf{70.43}                 & \textbf{38.84}        \\  \hline
\begin{tabular}[c]{@{}c@{}}NNMA\\ (three-level)\end{tabular} & 44.95           & \textbf{57.57}                             & \textbf{39.86}                 & 53.69        & 69.71                & 37.61       \\\hline
\end{tabular}
\caption{Performances of NNMA with Different Attention Levels. }
\label{tab:results1}
\end{table}
\begin{table*}[!htbp]
\centering
\begin{tabular}{|c|c|c|c|c|c|c|c|}
\hline
\multirow{2}{*}{System}       & \multicolumn{2}{c|}{Four-way}                                    & \multicolumn{5}{c|}{Binary}                                                                   \\ \cline{2-8}
                              & $F_1$                         & Acc.                                & Comp.                 & Cont.                 & Expa. & Expa.+EntRel                & Temp.                 \\ \hline
P\&C2012   & -                          & -                                   & 31.32                 & 49.82                 & -& 79.22                     & 26.57                 \\
J\&E2015 & -                          & -                                   & 35.93                 & 52.78                 & -&80.02                     & 27.63                 \\
Zhang2015 & 38.80                      & 55.39                               & 32.03                 & 47.08                & 68.96                &80.22 & 20.29                \\
R\&X2014            & 38.40                      & 55.50                               & 39.70                 & 54.40                 & 70.20  &80.44               & 28.70                 \\
R\&X2015  & 40.50                      & 57.10                               & \textbf{41.00}                 & 53.80                 & 69.40  &-               & 33.30                 \\ 
B\&D2015  & -                      & -                               & 36.36                &  55.76              & 61.76  &-               & 27.30                 
\\
Liu2016  &44.98                     & 57.27                              & 37.91                &  \textbf{55.88}                 & 69.97  &-               & 37.17                
\\
Ji2016  & 42.30                    & \textbf{59.50}                               & -                &  -                & -  &-               & -                 
\\ \hline
NNMA(two-level)               & \textbf{46.29}             & {57.17}                               & 36.70                 &54.48        & \textbf{70.43}     &80.73           & \textbf{38.84}        \\  
NNMA(three-level)               & 44.95           & 57.57                          & 39.86                 & 53.69       & 69.71                &\textbf{80.86}  &37.61       \\  \hline
\end{tabular}
\label{tab:results2}
\caption{Comparison with the State-of-the-art Approaches.}
\end{table*}
 
From Table~\ref{tab:results1}, we can see that the basic LSTM model  performs the worst.
With attention levels added, our NNMA model performs much better.
This confirms the observation above that one-pass reading is not enough for identifying the discourse relations.
With respect to the four-way $F_1$ measure, using NNMA with one-level attention produces a 4\% improvement over the baseline system with no attention.
Adding the second attention level gives another 2.8\% improvement.
We  perform significance test for these two improvements, and they are both significant under one-tailed t-test ($p<0.05$).
However, when adding the third attention level, the performance does not promote much and almost reaches its plateau.
We can see that three-level NNMA experiences a decease in $F_1$ and a slight increase in Accuracy compared to two-level NNMA.
The results imply that with more attention levels considered, our model 
may perform slightly better, but it  may incur the over-fitting problem due to adding more parameters.
With respect to the binary classification $F_1$ measures, we can see that the ``Comparison'' relation needs more passes of reading compared to the other three relations. The reason may be that the identification of the ``Comparison''  depends more on some deep analysis such as  semantic parsing, according to~\cite{zhou2010predicting}.

Next, we compare our models with six state-of-the-art baseline approaches, as shown  in Table~4.
The six baselines are introduced as follows.
\begin{itemize}
\setlength{\itemsep}{-2pt}
\item
\textbf{P\&C2012}: \newcite{park2012improving}  designed a  feature-based method and promoted the performance through optimizing the feature set.
\item
\textbf{J\&E2015}: \newcite{TACL536} used two recursive neural networks on the syntactic parse tree to induce the representation of the arguments  and  the entity spans.
\item
\textbf{Zhang2015}: \newcite{zhangshallow} proposed to use shallow convolutional neural networks to model  two arguments respectively. We replicated their model since they used a different setting in preprocessing  PDTB.
\item
\textbf{R\&X2014, R\&X2015}: \newcite{rutherford-xue:2014:EACL} selected lexical features, production rules, and Brown cluster pairs, and fed them into a maximum entropy classifier.
 \newcite{rutherfordimproving} further proposed to gather extra weakly labeled data based on the discourse connectives for the classifier.
 \item \textbf{B\&D2015}: \newcite{braud2015comparing}  combined several hand-crafted lexical features and word embeddings to train a max-entropy classifier.
  \item \textbf{Liu2016}: \newcite{liu2016multi} 
proposed to  better classify the discourse relations by learning from other discourse-related tasks with a multi-task neural network.
 \item \textbf{Ji2016}: \newcite{ji2016latent} proposed a neural language model over sequences of words and used the discourse relations as latent variables to connect the adjacent sequences.
\end{itemize}

It is noted that P\&C2012 and J\&E2015  merged the ``EntRel'' relation into the ``Expansion'' relation\footnote{EntRel is the entity-based coherence relation which is independent of implicit and explicit relations in PDTB. However some research merges it into the implicit Expansion relation. }.  
For a comprehensive comparison, we also experiment our model by adding a \textit{Expa.+EntRel vs Other} classification.
Our NNMA model with two attention levels exhibits obvious advantages over the six baseline methods on the whole.
It is worth noting that NNMA is even better than the  R\&X2015 approach which employs extra data. 

As for the performance on each discourse relation, with respect to the $F_1$ measure, we can see that our NNMA model can achieve the best results on the ``Expansion'', ``Expansion+EntRel'' and ``Temporal'' relations and competitive results on the  ``Contingency'' relation  .
The performance of recognizing  the  ``Comparison''  relation is only  worse than  R\&X2014 and R\&X2015.
As \cite{rutherford-xue:2014:EACL} stated, the  ``Comparison'' relation is closely related to the constituent parse feature of the text, like production rules. How to  represent and exploit these information in our model will be our  next research focus.
\begin{figure*}[!htbp]
	\centering
	\subfloat[Example with \textit{Comparison} relation]{
		\label{visual:subfig_a}
		\centering
		\includegraphics[height=1.05in]{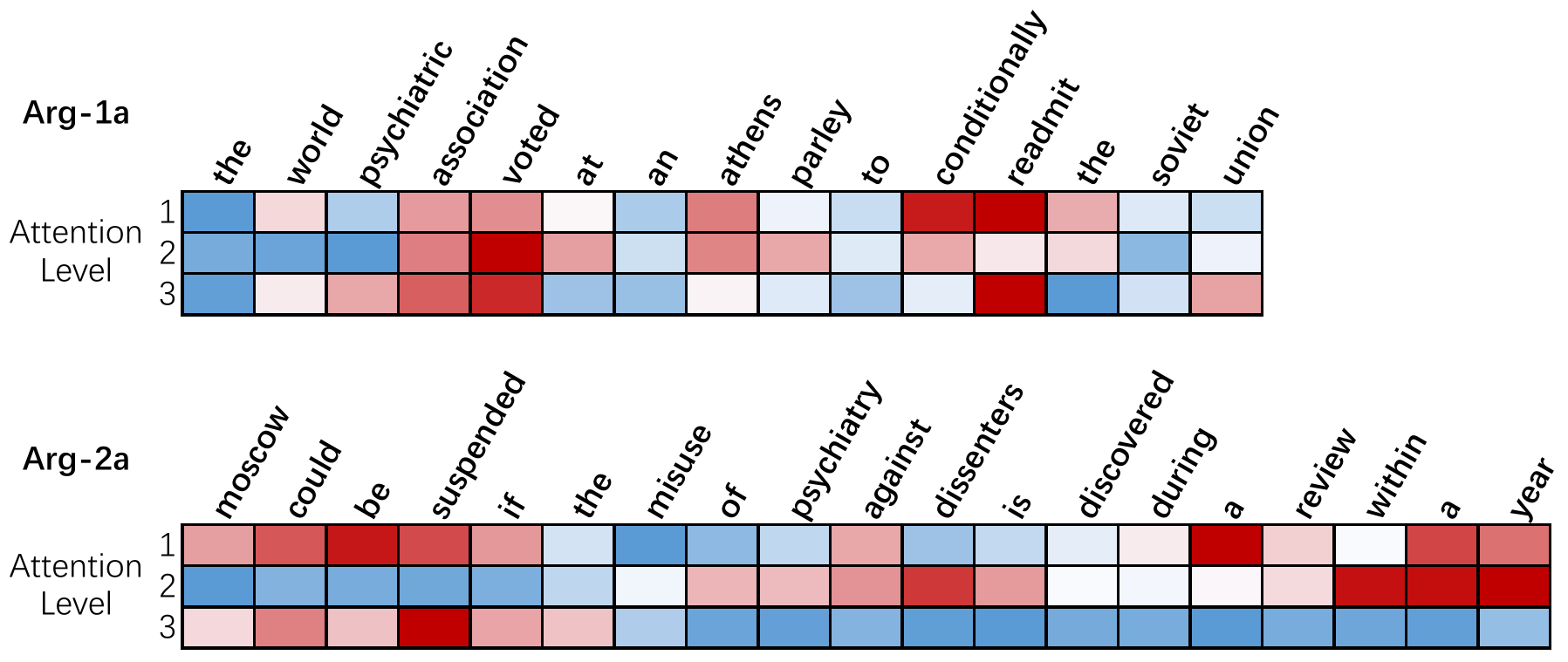}
	}
	\centering
	\subfloat[Example with \textit{Contingency} relation]{
		\label{visual:subfig_b}
		\centering
		\includegraphics[height=1.05in]{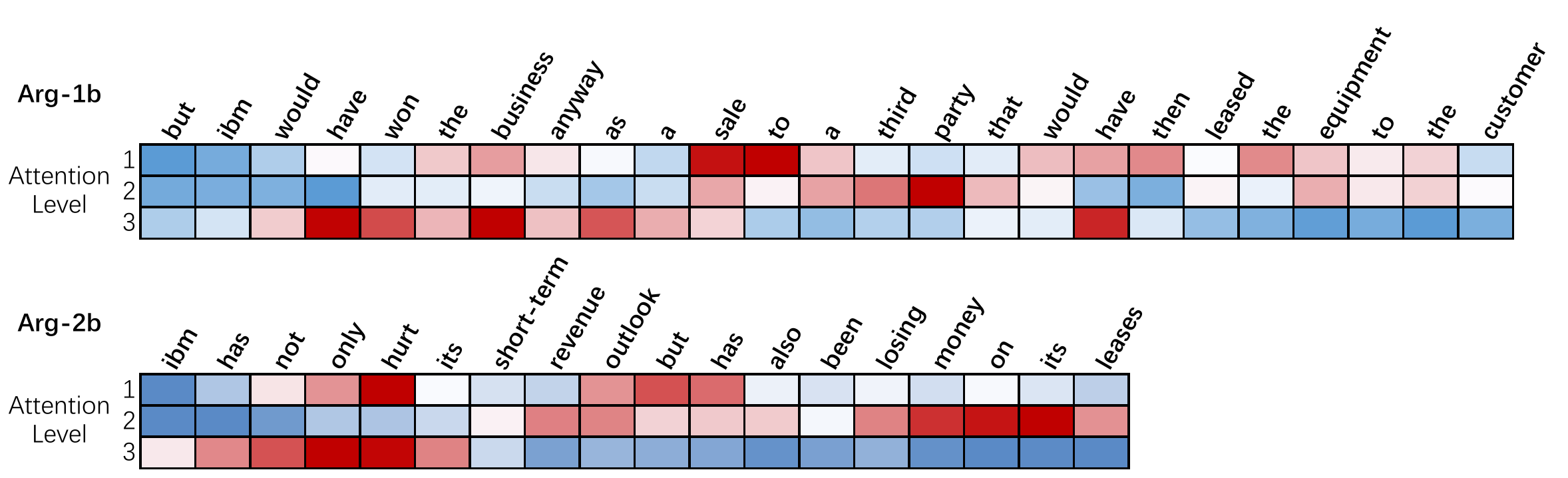}
	}
	\\
	\centering
	\subfloat[Example with \textit{Expansion} relation]{
		\label{visual:subfig_c}
		\centering
		\includegraphics[width=6.1in]{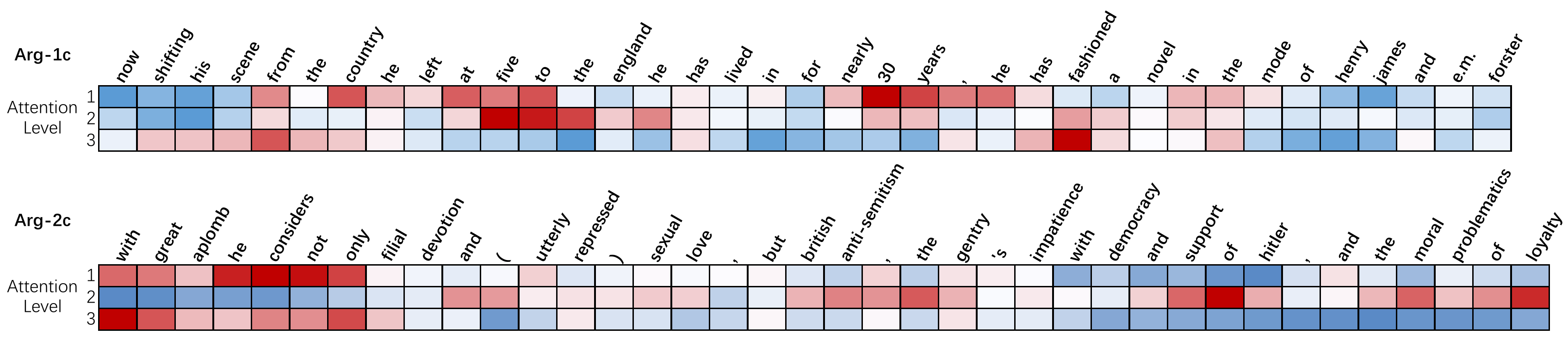}
	}
	\caption{ Visualization Examples: Illustrating Attentions Learned by NNMA. (The blue grid means the the attention on this word is lower than the value of a uniform distribution and the red red grid means the attention is higher than that.)}
	\label{visual}
\end{figure*}
\subsection{Analysis of Attention Levels}
The multiple attention levels in our model greatly boost the performance of classifying implicit discourse relations. In this subsection, we perform both  qualitative and quantitative analysis on the attention levels.

\begin{figure}[!htbp]
	\centering
	\includegraphics[height=2in]{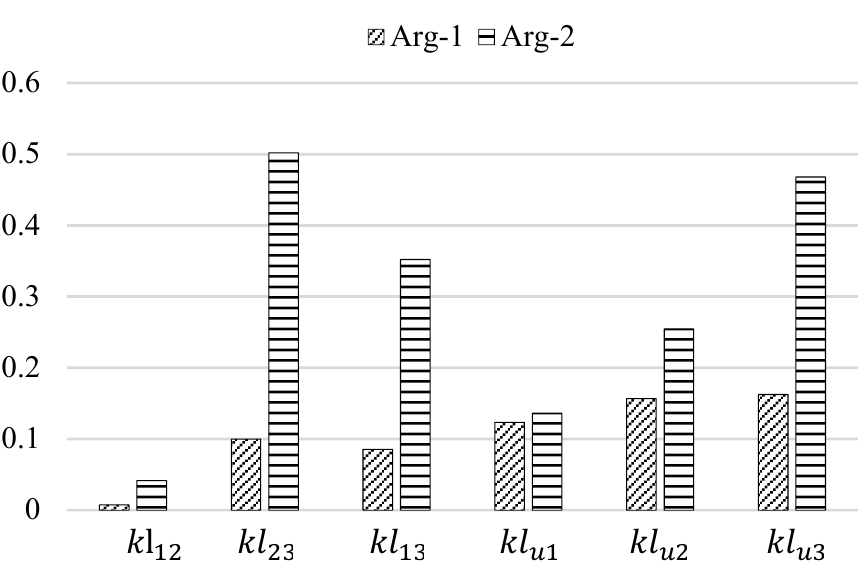}
	\caption{KL-divergences between attention levels}
	\label{fig:kl}
\end{figure}

First, we take a three-level NNMA model for example and analyze its attention distributions on different attention levels
by calculating the mean Kullback-Leibler (KL) Divergence between any two levels on the training set. In Figure~\ref{fig:kl}, we use $kl_{ij}$
to denote the KL Divergence between the $i^{th}$ and the $j^{th}$attention level  and use $kl_{ui}$  to denote the KL Divergence between the uniform distribution and the $i^{th}$ attention level. 
We can see that each attention level forms different attention distributions and the difference increases in the higher levels.  
It can be inferred that  the $2^{nd}$ and $3^{rd}$ levels in NNMA gradually neglect  some words and  pay more attention to some other words in the arguments.
One point worth mentioning is that \textit{Arg-2} tends to have more non-uniform attention weights, since $kl_{u2}$ and $kl_{u3}$ of \textit{Arg-2} are much larger than those of \textit{Arg-1}. 
And also, the changes between attention levels are more obvious for \textit{Arg-2} through observing the values of $kl_{12}$, $kl_{13}$ and $kl_{23}$.
The reason may be that  \textit{Arg-2}  contains more information related with discourse relation and some words in it tend to require focused attention, as \textit{Arg-2} is  syntactically bound to the implicit connective.

At the same time, we visualize the attention levels of some example argument pairs which are analyzed  by the three-level NNMA. 
To illustrate the $k^{th}$ attention level, we  get its attention weights $\bm{a}_k^1$ and $\bm{a}_k^2$ which reflect the contribution of each word  and then depict them by a row of color-shaded grids in Figure~\ref{visual}.
 
We can see that the NNMA model focuses on different  words on different attention levels.
Interestingly, from Figure~\ref{visual}, we find  that the $1^{st}$ and $3^{rd}$ attention levels focus on some similar words, while the $2^{nd}$ level is relatively different from them. 
It seems that NNMA tries to find some clues  (e.g. ``moscow could be suspended'' in \textit{Arg-2a}; ``won
the business'' in \textit{Arg-1b}; ``with great aplomb
he considers not only'' in \textit{Arg-2c}) for recognizing the discourse relation on the $1^{st}$ level,  looking closely at other words (e.g. ``misuse of psychiatry against dissenters'' in \textit{Arg-2a}; ``a third party that'' in \textit{Arg-1b}; ``and support of hitler'' in \textit{Arg-2c}) on the $2^{nd}$ level, and then reconsider the arguments, focus on some specific words (e.g. ``moscow could be suspended'' in \textit{Arg-2a}; ``has not only hurt'' in \textit{Arg-2b}) and make the final decision on the last level.

\section{Related Work}
\subsection{Implicit Discourse Relation Classification}
The Penn Discourse Treebank (PDTB)~\cite{10532931}, known as the largest discourse corpus,  is composed of 2159 Wall Street Journal articles. 
Each document is annotated with the predicate-argument structure, where the predicate is the discourse connective (e.g. while) and the arguments  are two text spans around the connective. The discourse connective can be  either explicit or implicit.
In PDTB, a hierarchy of relation tags is provided for annotation. In our study, we use the four top-level tags, including Temporal, Contingency, Comparison and Expansion. These four core relations allow us to be theory-neutral, since they are almost included in all discourse theories, sometimes with different names~\cite{wang2012coling}.

Implicit discourse relation recognition is often treated as a classification problem.
The first work to tackle this task on PDTB is~\cite{pitler2009automatic}. They selected several surface features to train four binary classifiers, each for one of the top-level PDTB relation classes.
Extending from this work, \newcite{lin2009recognizing} further identified four different feature types representing the context, the constituent parse trees, the dependency parse trees and the raw text respectively.
\newcite{rutherford-xue:2014:EACL}  used brown cluster to replace the word pair features for solving the sparsity problem.
\newcite{TACL536} adopted two recursive neural networks  to exploit the representation of arguments  and entity spans.
Very recently, \newcite{liu2016multi}  proposed a two-dimensional convolutional neural network (CNN) to model the argument pairs and employed a multi-task learning framework to boost the performance by learning from other discourse-related tasks.
\newcite{ji2016latent} considered discourse relations as latent variables connecting two token sequences and trained a discourse informed language model.

\subsection{Neural Networks and Attention Mechanism}
Recently, neural network-based methods have gained prominence in the field of natural language processing~\cite{kim:2014:EMNLP2014}.
Such methods are primarily based on learning a distributed representation for each word, which is also called a word embedding~\cite{collobert2011natural}.

Attention mechanism was first introduced into neural models to solve the alignment problem between different modalities. 
\newcite{graves2013generating}  designed a neural network to generate handwriting based on a text. It assigned a window on the input text at each step and generate characters based on the content within the window.  
\newcite{bahdanau2014neural}  introduced this idea into machine translation, where their model computed a probabilistic distribution over the input sequence when generating each target word.
 \newcite{tan2015lstm}   proposed an attention-based neural network to model both questions and sentences for selecting the appropriate non-factoid answers.

In parallel, the idea of equipping the neural model with an external memory has gained increasing attention recently. 
A memory can remember what the model has learned and guide its subsequent actions. 
\newcite{WestonCB14}  presented a  neural network to read and update the external memory in a recurrent manner with the guidance of a question embedding.
 \newcite{DBLP:journals/corr/KumarISBEPOGS15}  proposed a similar model where a memory was designed to change the gate of the gated recurrent unit for each iteration.

\section{Conclusion}
As a complex text processing task, implicit discourse relation recognition needs a deep analysis of the arguments.
To this end, we for the first time propose to imitate the repeated reading strategy and dynamically exploit efficient features  through several passes of reading. 
Following this idea, we design neural networks with multiple levels of attention (NNMA), where the general level and the attention levels  represent the first and subsequent passes of reading. 
With the help of external short-term memories, NNMA can gradually update the arguments representations on each attention level and  fix attention on some specific words which provide effective  clues to discourse relation recognition.
We conducted experiments on  PDTB and the evaluation results show that our model can achieve the state-of-the-art performance on recognizing the implicit discourse relations. 

\section*{ Acknowledgments}
We thank all the anonymous reviewers for their insightful comments on this paper. 
This work was partially supported by National Key Basic Research Program of China (2014CB340504), 
and National Natural Science Foundation of China (61273278 and 61572049). 
The correspondence author of this paper is Sujian Li.

\bibliography{emnlp2016}
\bibliographystyle{emnlp2016}
\end{document}